\documentclass[a4paper,conference]{IEEEtran}

\usepackage{amsmath}
\usepackage{graphicx}
\usepackage[breaklinks,colorlinks,pagebackref,urlcolor=blue]{hyperref}

\begin{document}

\title{A CNN Based Framework for Unistroke Numeral Recognition in Air-Writing}
\author{\IEEEauthorblockN{Prasun Roy, Subhankar Ghosh, and Umapada Pal}
\IEEEauthorblockA{Computer Vision and Pattern Recognition Unit, Indian Statistical Institute Kolkata, India}
\IEEEauthorblockA{\url{https://github.com/prasunroy/air-writing}}
}

\maketitle

\begin{abstract}
Air-writing refers to virtually writing linguistic characters through hand gestures in three-dimensional space with six degrees of freedom. This paper proposes a generic video camera-aided convolutional neural network (CNN) based air-writing framework. Gestures are performed using a marker of fixed color in front of a generic video camera, followed by color-based segmentation to identify the marker and track the trajectory of the marker tip. A pre-trained CNN is then used to classify the gesture. The recognition accuracy is further improved using transfer learning with the newly acquired data. The performance of the system varies significantly on the illumination condition due to color-based segmentation. In a less fluctuating illumination condition, the system is able to recognize isolated unistroke numerals of multiple languages. The proposed framework has achieved 97.7\%, 95.4\% and 93.7\% recognition rates in person independent evaluations on English, Bengali and Devanagari numerals, respectively.
\end{abstract}

\begin{IEEEkeywords}
Air-writing, human-computer interaction, gesture recognition, handwritten character recognition, convolutional neural networks.
\end{IEEEkeywords}

\section{Introduction}\label{sec:introduction}
Air-writing systems render a form of gestural human-computer interaction. Such systems are especially useful for building advanced user interfaces that do not require traditional mechanisms of linguistic input, such as pen-up-pen-down motion, hardware input devices or virtual keyboards. On the contrary, these advanced systems provide an interface for writing through hand gestures in three-dimensional space with six degrees of freedom. The input scheme in such systems fundamentally differs from a generic pen-up-pen-down input mechanism because, in the case of the former, there is no robust way of defining \emph{start} (pen-down) and \emph{stop} (pen-up) states while writing. Unlike conventional writing, air-writing systems lack actual anchoring and reference positions on the writing plane. Gestures are guided by considering imaginary axes in three-dimensional space. Consequently, these facts contribute to the increased variability of writing patterns for such systems, thereby accounting for the non-trivial nature of the problem.

The possibility of air-writing systems has emerged with the rapid development of depth sensors, such as Kinect \cite{kinect} and LEAP Motion \cite{leapmotion}, in recent years. Depth sensors and computer vision techniques are used to track fingertips, followed by recognition of the performed gestures using a trained model. However, these sensors are not widely available to common devices, which restricts these systems from being easily accessible. While depth sensors are not widely available, generic cameras are embedded in many commonly used devices. Therefore a generic video camera-aided air-writing system can be incredibly beneficial. However, unlike depth sensors, a generic camera does not provide information regarding scene depth or bone joints, making it more challenging to process and achieve reliable recognition accuracy. Common approaches for building air-writing systems involve various sensors, contributing to accurate motion tracking but limiting mass adoption with cost-effective general-purpose usage. In this work, attempts have been made to build a generic video camera-based air-writing system avoiding any other additional sensors.

The remainder of this paper is organized as follows. In Sec. \ref{sec:relatedwork}, some of the significant previous works are discussed. In Sec. \ref{sec:proposedwork}, the proposed work is presented. Sec. \ref{sec:results} describes the experimental results. Sec. \ref{sec:conclusion} concludes the paper with a summary of the work and potential scopes.

\begin{figure}[t]
    \centering
    \includegraphics[width=\linewidth]{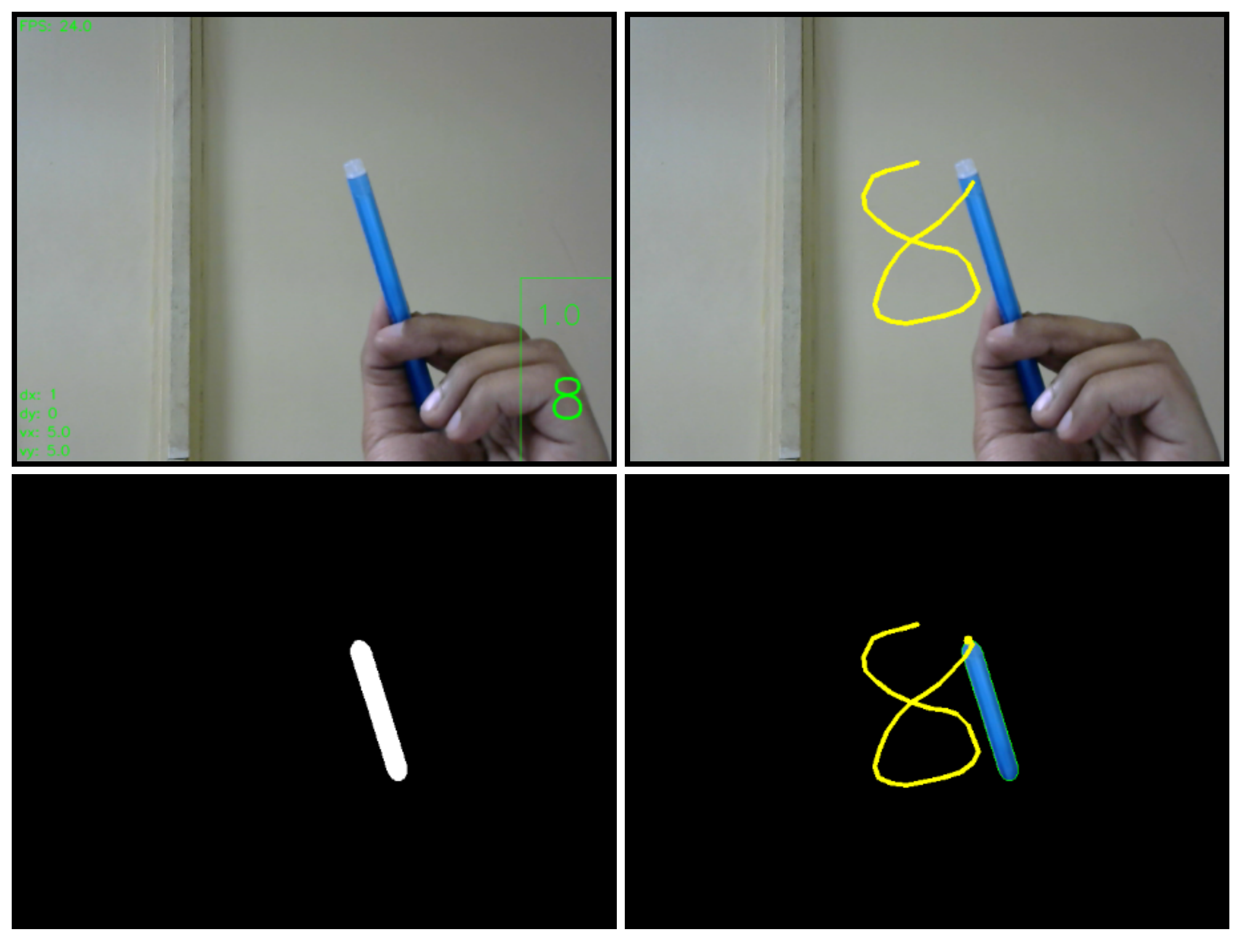}
    \caption{Overview of the proposed framework. \textbf{Top-left:} Original video frame. \textbf{Top-right:} Original video frame with approximate marker trajectory overlay. \textbf{Bottom-left:} Segmentation mask. \textbf{Bottom-right:} Segmented marker and approximate marker trajectory.}
    \label{fig:figure_1}
\end{figure}

\section{Related Work}\label{sec:relatedwork}
Most existing works on air-writing rely on depth sensors such as Kinect \cite{kinect} and LEAP Motion \cite{leapmotion}, or wearable gesture and motion control hardware such as Myo \cite{myo}. While these approaches offer highly accurate motion tracking that results in a better recognition rate, they limit the cost-efficient general adoption due to essential dependency on special-purpose external hardware.

Chen \textit{et al.} \cite{chen16p1, chen16p2} have used LEAP Motion for tracking and a Hidden Markov Model (HMM) for recognition that achieves 0.8\% error rate for word-based recognition and 1.9\% error rate for letter-based recognition. Kristensson \textit{et al.} \cite{kristensson12} have proposed a bimanual markerless interface for depth sensors using a probabilistic scale and translation invariant algorithm that achieves 92.7\% accuracy for one-handed and 96.2\% accuracy for two-handed gestures. Dash \textit{et al.} \cite{dash17} have utilized a Myo armband sensor along with a novel Fusion model architecture by combining one Convolutional Neural Network (CNN) and two Gated Recurrent Units (GRU). The Fusion model outperforms other widely used models such as HMM, SVM and KNN with an accuracy of 91.7\% in a person-independent evaluation and 96.7\% in a person-dependent evaluation. Schick \textit{et al.} \cite{schick12} have introduced a sensor-independent markerless framework using multiple cameras for 3D hand tracking, followed by recognition with HMM. This method achieves 86.15\% recognition rate for characters and 97.54\% recognition rate for isolated word recognition on a small vocabulary.

All previously proposed systems involve special-purpose sensors or a multi-camera setup which restricts mainstream adoption of these systems. This paper proposes a single camera-based air-writing framework that can be seamlessly integrated into many common devices with a built-in video camera, such as smartphones and laptops.

\section{Proposed Work}\label{sec:proposedwork}
\subsection{Marker segmentation}
Due to the high variability of human skin tone, it is significantly challenging to segment hands from the background by color-based segmentation technique. The proposed technique uses a marker object of fixed uniform color to mitigate the potential difficulties in hand and skin segmentation. Due to a uniform color distribution, the marker can be segmented from the background using a threshold. Assuming $f(x, y)$ and $g(x, y)$ to be pixel values at position $(x, y)$ of the initially captured video frame and segmented frame, respectively, and $I_m$ being the threshold for segmentation which is essentially the uniformly distributed pixel value of the marker object,

\begin{equation}
g(x, y) = 
\begin{cases}
1, & \text{if } f(x, y) = I_m\\
0, & \text{otherwise}
\end{cases}
\end{equation}

\subsection{Marker tip identification}
The resulting segmented binary image $g(x, y)$ contains the marker object and some noise. If the marker has a uniform distribution of color and this color is sufficiently distinctive from the background, the contour with the largest area in the segmented image can be labeled as the marker. The marker tip is estimated as the topmost point on the contour boundary with the lowest $y$-coordinate value.

\subsection{Trajectory approximation}
Unlike conventional \emph{pen-up-pen-down} motion-based writing with distinct breakpoints, the air-writing scheme is continuous. This fact contributes to higher complexity for segmenting air-written text into individual characters. This work proposes a velocity-based virtual \emph{pen-up-pen-down} motion estimation to address this difficulty. The effect of different rendering speeds of different cameras is normalized with the number of rendered frames per second ($N_{FPS}$) estimated as

\begin{equation}
N_{FPS} = \frac{1}{t_{update}}
\end{equation}
where $t_{update}$ is the time required to process the last video frame. Assuming $\Delta_x$ and $\Delta_y$ as changes in the position of the marker tip along $x$ and $y$ direction, respectively, between two consecutive frames, the normalized instantaneous velocity of the marker tip at time instance $t$ is estimated as

\begin{align}
dx	&= \frac{1}{N_{FPS}} \sum\limits_{t}^{t + N_{FPS}} \Delta_x\\
dy	&= \frac{1}{N_{FPS}} \sum\limits_{t}^{t + N_{FPS}} \Delta_y
\end{align}

\begin{figure}[ht]
    \centering
    \includegraphics[width=\linewidth]{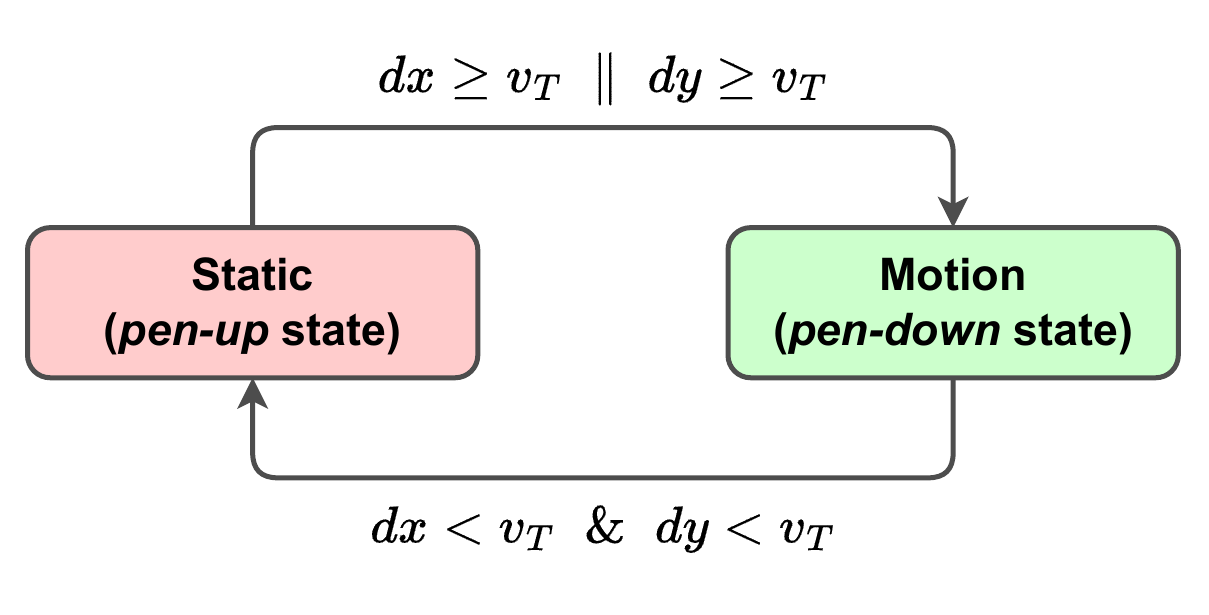}
    \caption{Velocity-based motion modelling scheme.}
    \label{fig:figure_2}
\end{figure}

The \emph{start} and \emph{stop} of continuous trajectory during air-writing is decided by comparing $dx$ and $dy$ with a velocity threshold $v_T$ estimated experimentally. When both $dx$ and $dy$ are below $v_T$, the marker is assumed to be \textbf{static} (\emph{pen-up} state). Otherwise, the marker is assumed to be in \textbf{motion} (\emph{pen-down} state). The trajectory of the marker tip is approximated as a piece-wise linear curve by considering straight line segments between marker tip positions in every two consecutive video frames when the marker is in the dynamic state. The motion modeling scheme is shown in Fig. \ref{fig:figure_2}.

\subsection{Character recognition}

\begin{figure*}[ht]
    \centering
    \includegraphics[width=\textwidth]{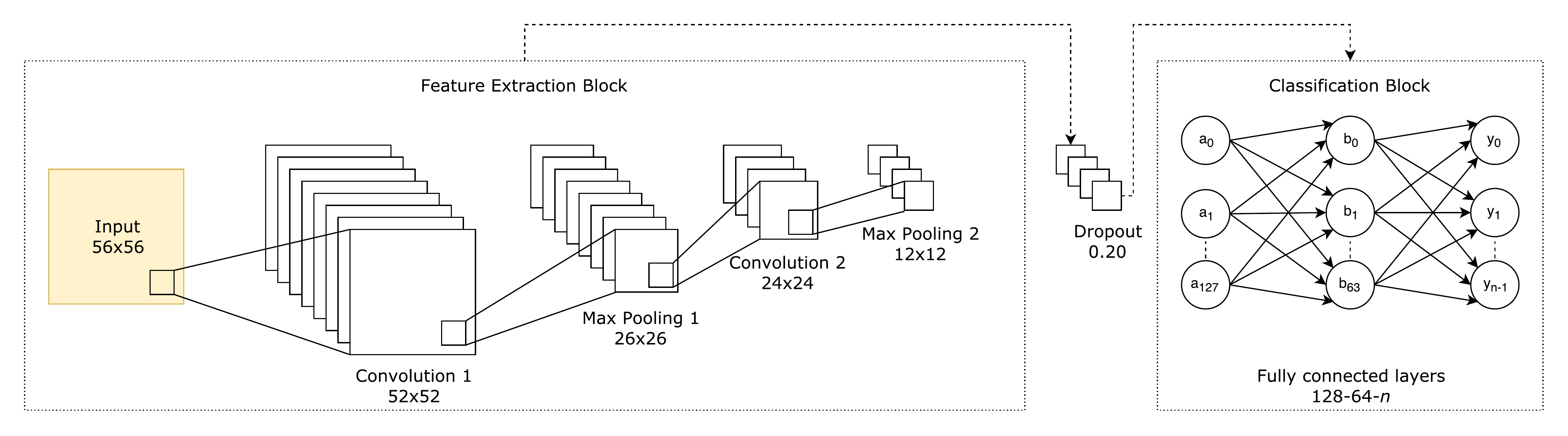}
    \caption{CNN architecture used in the proposed method.}
    \label{fig:figure_3}
\end{figure*}

The approximate trajectory of the marker tip is the projection of an air-written character from the three-dimensional space onto a two-dimensional image plane. A pre-trained convolutional neural network (CNN) is employed to predict the written character from this projected image. At the time of this study, no standard dataset is available for air-written numerals. Therefore the CNN model is initially trained on the handwritten digits of the standard MNIST dataset \cite{mnist98}. Later the pre-trained model is fine-tuned on a smaller dataset of newly acquired air-written characters.

 Fig. \ref{fig:figure_3} shows the architecture of the proposed CNN. The model includes a feature extraction block followed by a classification block. The feature extraction block takes a $56 \times 56$ grayscale image as input, and it consists of two convolution layers, each followed by a pooling layer. The first convolution layer uses 32 filters and a $5 \times 5$ kernel. The second convolution layer uses 16 filters and a $3 \times 3$ kernel. Both convolution layers use rectified linear units (ReLU) as activation functions. The classification block consists of three fully connected layers with 128, 64 and $n$ computing units (neurons), respectively, where $n$ is the number of output classes corresponding to the character set under consideration. The first two fully connected layers use ReLU as the activation function, while the final layer uses a normalized exponential function (softmax) for the same purpose. The possibility of overfitting is addressed using 20\% dropout between the two blocks.

\section{Results \& Discussion}\label{sec:results}
\subsection{Data Acquisition}
To the best of our knowledge, no standard dataset of air-written characters is currently available for benchmarking the proposed method. For this reason, a dataset of air-written numerals is compiled to aid further experimentation. Data is recorded using a marker of fixed uniform color and a generic video camera. After marker segmentation, marker tip identification, followed by trajectory approximation, an image of the locus of the marker tip is obtained. Each resulting image instance is then numerically labeled and appended to the dataset. Three different datasets for English, Bengali and Devanagari air-written numerals are separately prepared. Each dataset contains 10000 air-written numerals collected from 20 individuals, each writing one numeral 50 times. For pre-training, three standard datasets of handwritten numerals, one for each language, are prepared. These datasets consist of 70000 English handwritten numerals from the MNIST dataset \cite{mnist98}, 14650 Bengali and 22546 Devanagari handwritten numerals from \cite{pal07}. The air-writing dataset of each language is divided into two disjoint sets -- a training set (TS-A) with 6000 instances and a test set (EVAL) with 4000 instances. The other training set (TS-B) includes 70000 English, 14650 Bengali and 22546 Devanagari handwritten numerals. Dataset distributions of English, Bengali and Devanagari numerals are shown in Table \ref{tab:table_1}, \ref{tab:table_2} and \ref{tab:table_3}, respectively.

\begin{table}[h]
\centering
\caption{Dataset distribution for English numerals.}
\label{tab:table_1}
\begin{tabular}{l|c|c|c}
\hline
\textbf{Dataset} & \textbf{Type} & \textbf{Source dataset} & \textbf{\#Instance} \\
\hline \hline
TS-A    & Training Set A        & Air-Writing          & ~6000 \\
TS-B    & Training Set B        & MNIST \cite{mnist98} & 70000 \\
EVAL    & Test Set              & Air-Writing          & ~4000 \\
\hline
\end{tabular}
\end{table}

\begin{table}[h]
\centering
\caption{Dataset distribution for Bengali numerals.}
\label{tab:table_2}
\begin{tabular}{l|c|c|c}
\hline
\textbf{Dataset} & \textbf{Type} & \textbf{Source dataset} & \textbf{\#Instance} \\
\hline \hline
TS-A    & Training Set A        & Air-Writing          & ~6000 \\
TS-B    & Training Set B        & Bengali \cite{pal07} & 14650 \\
EVAL    & Test Set              & Air-Writing          & ~4000 \\
\hline
\end{tabular}
\end{table}

\begin{table}[h]
\centering
\caption{Dataset distribution for Devanagari numerals.}
\label{tab:table_3}
\begin{tabular}{l|c|c|c}
\hline
\textbf{Dataset} & \textbf{Type} & \textbf{Source dataset} & \textbf{\#Instance} \\
\hline \hline
TS-A    & Training Set A        & Air-Writing             & ~6000 \\
TS-B    & Training Set B        & Devanagari \cite{pal07} & 22546 \\
EVAL    & Test Set              & Air-Writing             & ~4000 \\
\hline
\end{tabular}
\end{table}

\subsection{Experimental setup}
The model is trained and evaluated on different combinations of TS-A, TS-B and EVAL splits as follows.

\begin{enumerate}
    \item Training with TS-A and testing on EVAL.
    \item Training with TS-B and testing on EVAL.
    \item Training with TS-A + TS-B combined and testing on EVAL.
    \item Training with TS-B followed by fine-tuning with TS-A and testing on EVAL
\end{enumerate}

\begin{figure*}[t]
    \centering
    \includegraphics[width=\linewidth]{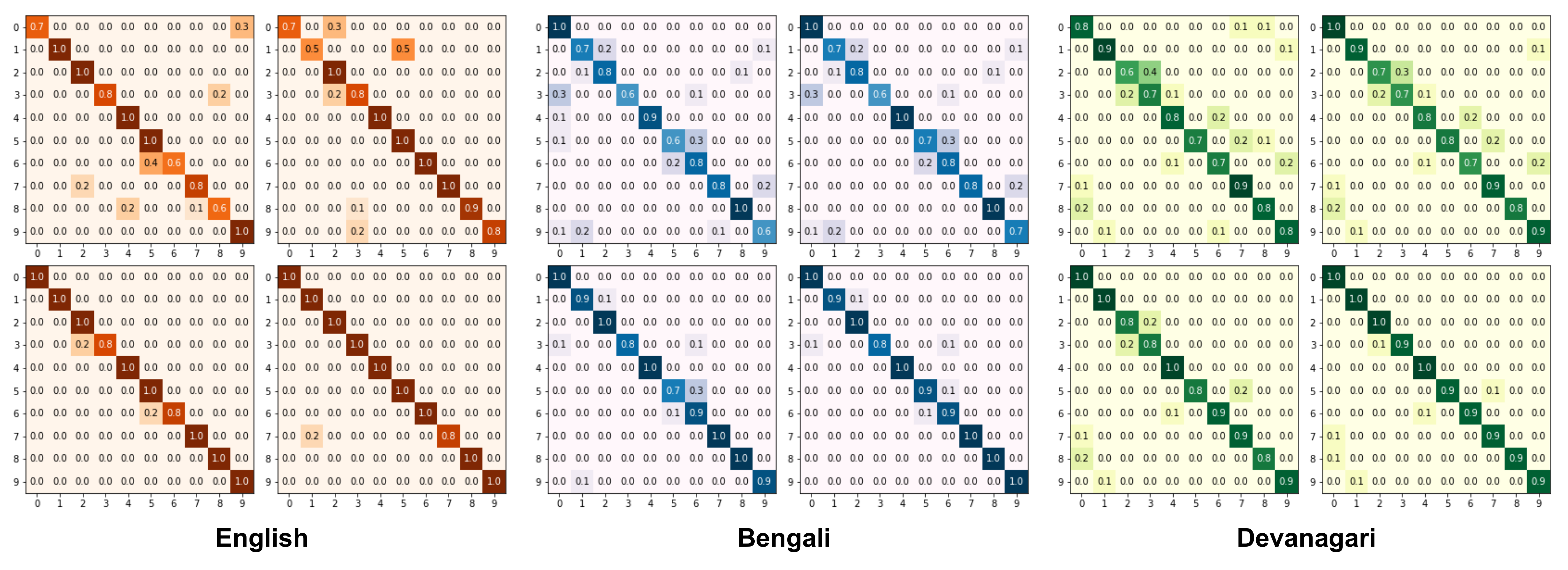}
    \caption{Confusion matrices for English, Bengali and Devanagari numerals. For each language -- {\textbf{Top-left:} Training on TS-A and testing on EVAL. \textbf{Top-right:} Training on TS-B and testing on EVAL. \textbf{Bottom-left:} Training on TS-A + TS-B combined and testing on EVAL. \textbf{Bottom-right:} Training on TS-B followed by fine-tuning on TS-A and testing on EVAL.}}
    \label{fig:figure_4}
\end{figure*}

\begin{table*}[t]
\centering
\caption{Test accuracy for English, Bengali and Devanagari numerals.}
\label{tab:table_4}
\resizebox{0.95\textwidth}{!}{%
\begin{tabular}{l|c|c|c|c|c}
\hline
\textbf{Training set} & \textbf{Fine-tuning set} & \textbf{Test set} & \textbf{Accuracy (English)} $\uparrow$ & \textbf{Accuracy (Bengali)} $\uparrow$ & \textbf{Accuracy (Devanagari)} $\uparrow$ \\
\hline \hline
TS-A        & ~--~ & EVAL & 81.8\% & 78.1\% & 77.3\% \\
TS-B        & ~--~ & EVAL & 86.4\% & 81.1\% & 82.4\% \\
TS-A + TS-B & ~--~ & EVAL & 95.5\% & 92.5\% & 89.5\% \\
TS-B        & TS-A & EVAL & \textbf{97.7}\% & \textbf{95.4}\% & \textbf{93.7}\% \\
\hline
\end{tabular}
}
\end{table*}

\begin{table*}[t]
\centering
\caption{Examples of misclassified samples.}
\label{tab:table_5}
\resizebox{0.95\textwidth}{!}{%
\begin{tabular}{|c|c|c|c||c|c|c|c|}
\hline
\textbf{Sample} & \textbf{Actual class} & \textbf{Predicted class} & \textbf{Confidence} & \textbf{Sample} & \textbf{Actual class} & \textbf{Predicted class} & \textbf{Confidence} \\
\hline \hline
\includegraphics[scale=0.5]{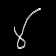}
&
\includegraphics[scale=0.5]{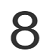}
&
\includegraphics[scale=0.5]{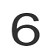}
&
51.5\%
&
\includegraphics[scale=0.5]{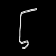}
&
\includegraphics[scale=0.5]{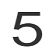}
&
\includegraphics[scale=0.5]{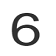}
&
49.8\% \\ \hline
\includegraphics[scale=0.5]{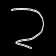}
&
\includegraphics[scale=0.5]{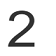}
&
\includegraphics[scale=0.5]{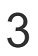}
&
37.1\%
&
\includegraphics[scale=0.5]{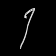}
&
\includegraphics[scale=0.5]{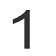}
&
\includegraphics[scale=0.5]{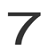}
&
44.9\% \\ \hline
\includegraphics[scale=0.5]{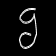}
&
\includegraphics[scale=0.5]{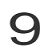}
&
\includegraphics[scale=0.5]{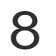}
&
55.2\%
&
\includegraphics[scale=0.5]{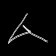}
&
\includegraphics[scale=0.5]{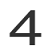}
&
\includegraphics[scale=0.5]{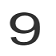}
&
23.2\% \\ \hline
\includegraphics[scale=0.5]{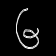}
&
\includegraphics[scale=0.5]{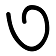}
&
\includegraphics[scale=0.5]{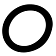}
&
19.5\%
&
\includegraphics[scale=0.5]{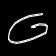}
&
\includegraphics[scale=0.5]{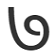}
&
\includegraphics[scale=0.5]{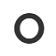}
&
57.4\% \\
\hline
\end{tabular}
}
\end{table*}

\subsection{Results}
Table \ref{tab:table_4} shows the quantitative comparison of test accuracy on EVAL for different combinations of training sets TS-A and TS-B for each language. In every case, the best performance is obtained while pre-training on handwritten samples, followed by fine-tuning with air-written samples. The proposed framework is not directly comparable with most of the previous works because of the difference in evaluation methods and types of datasets. However, under comparable evaluation conditions, the proposed framework achieved a 6\% better recognition rate than a recently proposed method \cite{dash17}. Our proposed system achieved 97.7\% accuracy for air-written English numerals whereas Dash \textit{et al.} \cite{dash17} obtained 91.7\% accuracy using person independent evaluation protocol.

\subsection{Error analysis}
Figure. \ref{fig:figure_4} shows the confusion matrices obtained during evaluation on EVAL with each combination of TS-A and TS-B for English, Bengali and Devanagari numerals. The recognition rate is significantly improved when pre-training the CNN with a larger dataset of handwritten numerals, followed by fine-tuning the model with a much smaller dataset of air-written numerals. This can be achieved due to geometric similarities between handwritten and air-written numerals. Upon pre-training on a larger dataset of handwritten numerals, the convolution layers are initially trained to extract features. Afterward, the model is fine-tuned on a much smaller dataset of air-written numerals to achieve a robust recognition performance through domain adaptation.

The proposed technique has achieved a reasonably robust recognition performance on numerals of multiple languages in real-time tests. However, there are certain occasions when the model fails to predict the air-written characters correctly, potentially due to malformed characters during trajectory tracing caused by unintended movements (shakes) of the camera or user. Table \ref{tab:table_5} shows a few examples of misclassified numerals recorded during real-time testing, along with actual class (intended by the user), predicted class (inferred by the model) and corresponding confidence scores.

\section{Conclusion}\label{sec:conclusion}
This paper proposes a robust framework for multi-language unistroke air-written numeral recognition. To avoid the difficulties of human skin segmentation, a marker of uniform color is used. Recognition is performed by a CNN, pre-trained on a large dataset of handwritten numerals, followed by domain adaptation through fine-tuning on a small dataset of air-written numerals. In experiments, the proposed framework has achieved 97.7\%, 95.4\% and 93.7\% recognition rates over English, Bengali and Devanagari numerals, respectively. The primary advantage of the method is a cost-efficient and easily adaptable approach that is entirely independent of any depth or motion sensors, such as Kinect, LEAP Motion and Myo Armband. The framework can be seamlessly integrated into any common device having a generic video camera. Adopting the framework directly for hands without requiring a fixed marker is the potential scope of a major improvement of the method for better flexibility in general usage.

\bibliographystyle{IEEEtran}
\bibliography{references}

\end{document}